\documentclass[10pt,twocolumn,letterpaper]{article}

\usepackage{wacv}              

\usepackage{graphicx}
\usepackage{amsmath}
\usepackage{amssymb}
\usepackage{booktabs}
\usepackage{pifont}

\usepackage[pagebackref,breaklinks,colorlinks]{hyperref}

\usepackage[capitalize]{cleveref}
\crefname{section}{Sec.}{Secs.}
\Crefname{section}{Section}{Sections}
\Crefname{table}{Table}{Tables}
\crefname{table}{Tab.}{Tabs.}

\usepackage{soul}
\usepackage{xcolor}

\def\wacvPaperID{229} 
\def\confName{WACV}
\def\confYear{2025}

\begin{document}

\title{GaussianBeV : 3D Gaussian Representation meets Perception Models for BeV Segmentation}

\author{Florian Chabot
\and
Nicolas Granger
\and
Guillaume Lapouge \\
\and
CEA, List, F-91120, Palaiseau, France\\
{\tt\small \string{firstname.lastname\string}@cea.fr}
}
\maketitle

\begin{abstract}
The Bird’s-eye View (BeV) representation is widely used for 3D perception from multi-view camera images. It allows to merge features from different cameras into a common space, providing a unified representation of the 3D scene. The key component is the view transformer, which transforms image views into the BeV. However, actual view transformer methods based on geometry or cross-attention do not provide a sufficiently detailed representation of the scene, as they use a sub-sampling of the 3D space that is non-optimal for modeling the fine structures of the environment. In this paper, we propose GaussianBeV, a novel method for transforming image features to BeV by finely representing the scene using a set of 3D gaussians located and oriented in 3D space. This representation is then splattered to produce the BeV feature map by adapting recent advances in 3D representation rendering based on gaussian splatting \cite{gs}. GaussianBeV is the first approach to use this 3D gaussian modeling and 3D scene rendering process in an optimization free manner, i.e. without optimizing it on a specific scene and directly integrated into a single stage model for BeV scene understanding. Experiments show that the proposed representation is highly effective and place GaussianBeV as the new state-of-the-art on the BeV semantic segmentation task on the nuScenes dataset \cite{nuscenes}.
\end{abstract}

\section{Introduction}
\label{sec:intro}

\begin{figure}[t]
    \centering
    \begin{subfigure}[b]{0.225\textwidth}
        \centering
        \includegraphics[width=\textwidth]{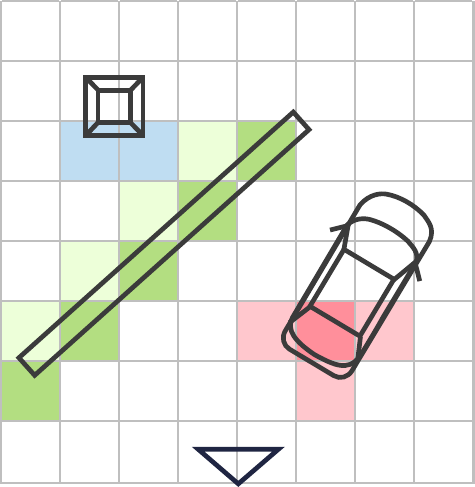}
        \caption{Depth-based \cite{lss,fiery,bevdet,bevdepth}}
        \label{fig:methods_overview_depth}
    \end{subfigure}
    \hfill
    \begin{subfigure}[b]{0.225\textwidth}
        \centering
        \includegraphics[width=\textwidth]{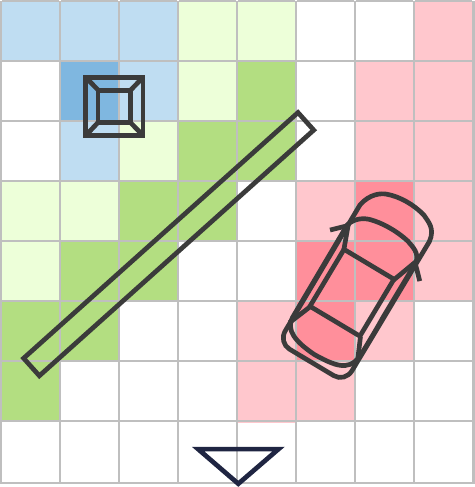}
        \caption{Projection-based \cite{simplebev, pointbev}}
        \label{fig:methods_overview_projection}
    \end{subfigure}
    
    \vfill
    \begin{subfigure}[b]{0.225\textwidth}
        \centering
        \includegraphics[width=\textwidth]{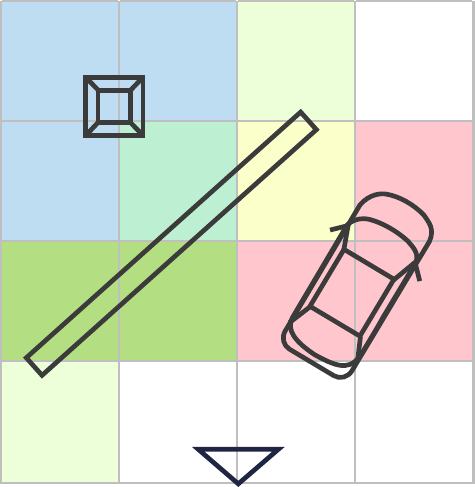}
        \caption{Attention-based \cite{cvt, petrv2} }
        \label{fig:methods_overview_attention}
    \end{subfigure}
    \hfill
    \begin{subfigure}[b]{0.225\textwidth}
        \centering
        \includegraphics[width=\textwidth]{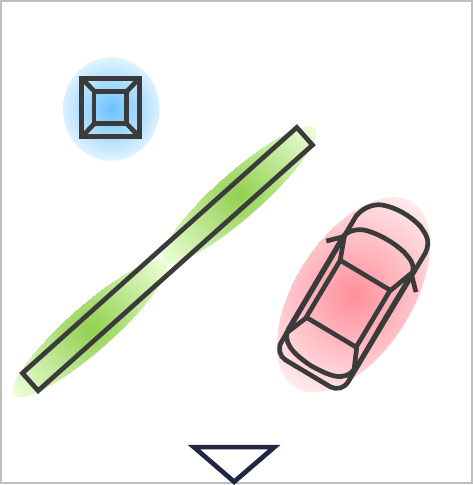}
        \caption{GaussianBeV (ours)}
        \label{fig:methods_overview_GS}
    \end{subfigure}
    \caption{Illustration of multiple BeV representations for BeV semantic segmentation. A camera is represented by the triangle at the bottom of each BeV. Features are represented by colors where blue, red and green represent the streetlight, car and lane marking respectively. (a) Depth-based methods place image features along the optical ray on the surface of objects. (b) In projection-based methods, 3D points on the optical ray receive the same feature. (c) Attention-based methods use downsampled dense spatial queries to  keep memory costs down. (d) In GaussianBeV, the scene is represented by a set of rotated gaussians that finely describes the semantic structures in the scene. }
    \label{fig:intuition}
\end{figure}

Multi-camera 3D perception tasks, such as semantic segmentation, are crucial for autonomous navigation applications. A common strategy involves projecting and merging features from different cameras into a bird’s-eye view (BeV) representation, which is then analyzed by perception heads. The primary challenge in these approaches lies in addressing the loss of 3D information during the projection of the physical world into camera images, thus solving the inverse problem of transforming image views into the BeV.

Recent literature identifies three main subsets of methods for image-to-BeV transformation. First, depth-based methods \cite{lss,fiery,bevdet,bevdepth} achieve view transformation geometrically by filling a 3D grid with features extracted from images based on the prediction of discrete depth distribution. The key idea is to roughly localize the 3D position of each image feature and then accumulate them through a voxelization step. However, in these approaches, 3D feature localization depends on the depth discretization granularity and is sub-optimal, as features are placed at the level of the visible faces of objects. Second, projection-based methods \cite{simplebev, pointbev} also use a geometric approach, project 3D grid points into the cameras and gathers corresponding features from them. While straightforward, these methods do not yield accurate 2D-to-3D back-projection, as all grid points along the same camera ray receive the same feature.  Third, transformer-based methods \cite{bevformer, cvt, petrv2} utilize cross-attention to merge multi-view features. Though effective for 3D object detection, their application to dense tasks like BeV semantic segmentation incurs a high computational cost due to dense spatial queries needed for BeV representation in the attention process. Some works \cite{cvt, petrv2} address this by reducing the BeV resolution, leading to inevitable information loss.

In this article, we propose a novel view transformation method called GaussianBeV, which enables fine 3D modeling of scenes. Drawing on recent advances in explicit 3D representation for rendering novel views based on Gaussian Splatting (GS) \cite{gs}, our method represents a scene using a set of 3D gaussians, each parameterized by a center, scale, rotation, opacity and \textit{semantic features} (instead of colors in GS). Furthermore, unlike the original GS method, which uses offline optimization for a specific scene to determine the 3D gaussian representation, we propose to train a neural network to directly generate an \textit{optimization free} 3D gaussian representation of the scene from multi-view images. This representation is then rendered into a BeV feature map which is analyzed by semantic segmentation heads. 

The representation of a scene by a set of 3D gaussians allows to model its entire content. Indeed, the geometrical properties of the gaussians (position, size and rotation) enable to cover 3D space with varying level of detail depending on the structures encountered in the scene. Intuitively, a gaussian representing a lane marking will be rotated and elongated along its length. A gaussian representing a vehicle will be placed in the center of the vehicle and will follow its shape. Figure \ref{fig:methods_overview_GS} illustrates the intuition behind the representation proposed in this paper. Our contributions can be summarized as follows. (1) Introduction of GaussianBeV for BeV feature map generation from images through an \textit{optimization free} image-to-3D gaussian representation for any scene, allowing fine 3D content modeling. This representation is then splattered in BeV using a rasterizer module. To our knowledge, this is the first time that a gaussian splatting representation that is not scene-specific is proposed and integrated into a BeV perception model. (2) Experiments demonstrating the effectiveness of our method, establishing it as the new state-of-the-art in BeV semantic segmentation.

\begin{figure*}[t]
  \centering
  \includegraphics[width=1.0\linewidth]{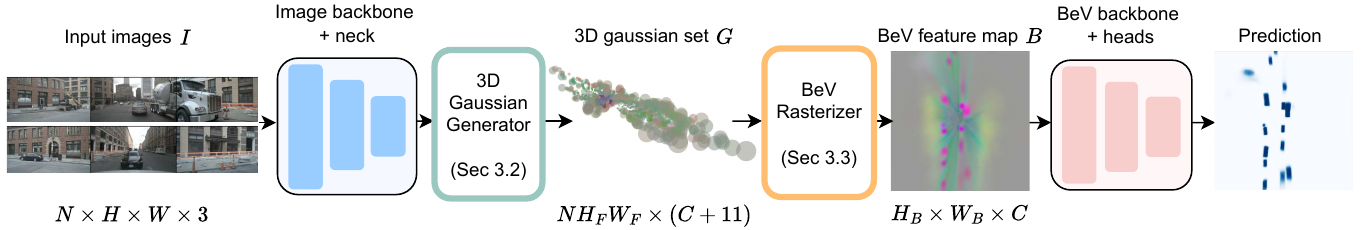}
  \caption{\textbf{Overview of GaussianBeV}. The network takes as input a set of multiview images and extracts features for each of them. The 3D gaussian generator module (Sec \ref{sec:gaussiangenerator}) predicts a 3D gaussian representation $G$ of the scene which is then sent to the BeV rasterizer module (Sec \ref{sec:bevrasterizer}) to performs BeV rendering. The resulting BeV feature map $B$ is passed through a BeV backbone and segmentation heads to obtain the segmentation prediction. $G$ and $B$ are represented with colors only for visualization purpose.}
  \label{fig:schema}
\end{figure*}

\section{Related work}
\label{sec:sota}

\paragraph*{Depth-based.}
A series of models were built on the explicit prediction of pixel-wise depth estimations along with image features. Combined with camera calibration parameters, this enables the back-projection of 2D features into 3D feature point cloud which is finally aggregated in the BeV grid. To accommodate for uncertainty in depth estimation, the features are actually propagated all along the ray that traverses their pixel and modulated by a discrete depth probability estimation~\cite{lss,fiery,bevdet}. To improves depth prediction, an explicit depth supervision scheme is proposed~\cite{bevdepth} using LiDAR data during model training.
However, depth-based methods are sensitive to the ray sampling strategy, usually back-projecting features along the ray and on object surfaces (see Figure \ref{fig:methods_overview_depth}). \\

\noindent \textbf{Projection-based.}
Using a thorough comparative study, \cite{simplebev} proposes to discard depth estimation in favor of a simpler projection scheme: a predefined set of 3D points is used to describe the scene and camera features are probed by projecting the point on the camera feature maps using calibration data.
This projection disregards actual objects and background placement but returns a denser representation of the scene without void beyond the depth of objects surfaces.
The computational and memory overhead of the generating the BeV grid is reduced by opting for a sparse grid representation~\cite{pointbev}. Projection-based view transformation methods are simple but result in a coarse BeV representation because all voxels along the optical ray receive the same features (see Figure \ref{fig:methods_overview_projection}). \\
\noindent \textbf{Attention-based.} Capitalising on recent advances in Transformer models, depth estimation is replaced by an attention-based feature modulation scheme~\cite{bevformer, petrv2, cvt, sparsebev, streampetr}.
Several optimization schemes are proposed to resolve computational complexity of a pairwise matching between image and BeV grid tokens: factorization of spatial and temporal attentions, deformable attention~\cite{bevformer}, injection of calibration and timestamp priors~\cite{petrv2}. 
For the segmentation task, attention-based view transformation is computationally and memory intensive, due to the need to define a dense query map \cite{bevformer}. This is why some methods \cite{cvt, petrv2} predict a low-resolution BeV (see Figure \ref{fig:methods_overview_attention}), which is then upsampled by successive deconvolutions.\\
\noindent \textbf{Gaussian splatting.} Gaussian splatting (GS)~\cite{gs} is a 3D scene rendering technique which uses 3D gaussians to describe a scene.
Each gaussian is parameterized by its position, scale, rotation, opacity and a Spherical Harmonics color model.
The entire rendering pipeline is differentiable allowing the optimization of the gaussian parameters to a particular scene based on a set of images. GS is both fast and parallel, allowing real-time operation on GPUs.

In comparison to sparse voxel grids, gaussians offer more efficient representations of a scene since individual gaussians can describe large volumes while smaller ones can accurately encode finer details with arbitrary resolution. 
Several extensions have been proposed, allowing the management of dynamic objects \cite{spacetimegaus, streetgaussians} or the distillation of semantic features from foundation models in the representation \cite{shi2024language,langsplat, zhou2024feature,ye24gaussiangrouping}.
Although GS rasterization itself is fast, the optimization of gaussian parameters to match a scene usually requires many iterations.
In order to address this issue for real-time SLAM, \cite{matsuki2024gaussiansplattingslam} leverages ground-truth 3D information from a depth sensor to initialize gaussian positions and \cite{yan2023gs} uses temporal consistency in order to minimize the optimization workload between successive frames.
To perform scene reconstruction with stereo images \cite{charatan23pixelsplat} or object-centric single image \cite{ szymanowicz24splatter}, optimization-free methods are introduced allowing the model to directly learn to predict the gaussian parameters.

In our work, we propose to overcome the drawbacks of view transformer methods by modelling scenes as set of gaussians which are subsequently rasterized into a feature map.
Unlike previous gaussian splatting works which optimize the gaussian representations to each scene, we propose to learn a neural network capable of directly predicting a gaussian representation of any scene.
Compared to~\cite{charatan23pixelsplat, szymanowicz24splatter} which targets scene reconstruction, our model is designed to perform real-time outdoor BeV semantic segmentation from multiple calibrated cameras.

\section{GaussianBeV}
\label{sec:method}

\subsection{Overview}

\begin{figure*}[ht]
  \centering
  \includegraphics[width=0.99\linewidth]{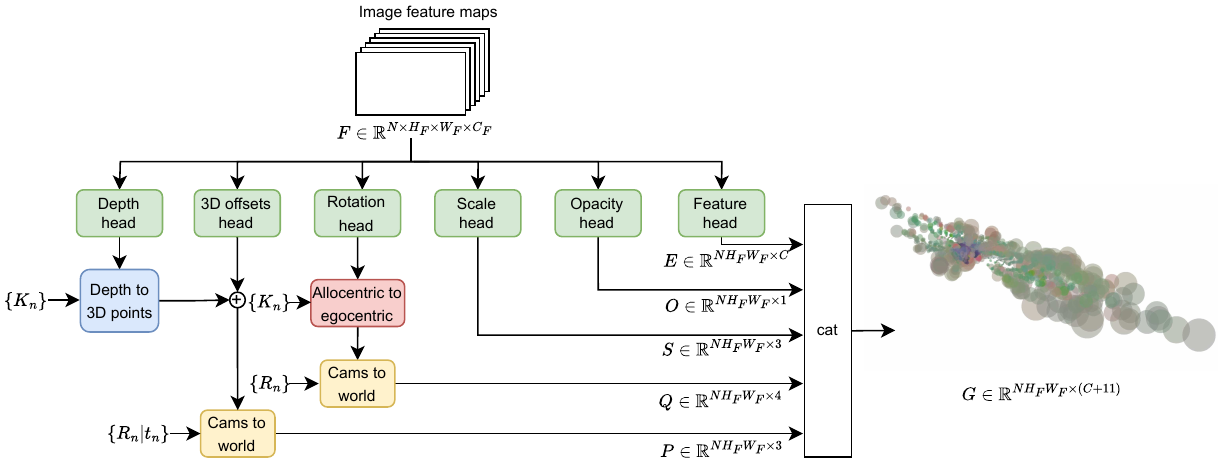}
  \caption{\textbf{3D Gaussian generator}. This module takes as input the feature map extracted from each camera, as well as the set of intrinsic $\left\{K_n\right\}$ and extrinsic $\left\{R_n| t_n\right\}$ parameters. For each pixel, it calculates the corresponding 3D gaussian by passing through prediction heads (green boxes). Some of these predictions are then decoded (blue and red boxes) and transformed to be expressed in the world reference frame (yellow boxes). All predicted gaussian parameters are then concatenated to produce the 3D gaussian representation $G$ of the scene.
}
  \label{fig:gg}
\end{figure*}

Figure \ref{fig:schema} presents an overview of GaussianBeV. The model takes as input a set of multiview images $I \in \mathbb{R}^{N \times H \times W \times 3}$ with $N$ the number of cameras, $H$ and $W$ the dimensions of the images. These images are passed sequentially through four modules, leading to BeV segmentation. 

The first module extracts image features using an image backbone and a neck to obtain feature maps $F \in \mathbb{R}^{N \times H_F \times W_F \times C_F}$, with $C_F$ the number of channels, $H_F$ and $W_F$ the dimensions of the feature maps. 

The second module is the 3D gaussian generator (Sec~\ref{sec:gaussiangenerator}) that predicts for each pixel in the feature maps the parameters of the corresponding gaussian in the world reference frame.
The output of this module is a set of 3D gaussians $G \in \mathbb{R}^{N H_F W_F \times (C + 11)}$ with $C$ the number of channels of the embedding associated to each gaussian.
More specifically, $G$ contains the following parameters : positions $P \in \mathbb{R}^{N H_F W_F \times 3}$, scales $S \in \mathbb{R}^{N H_F W_F \times 3}$, rotations as unit quaternions $Q \in \mathbb{R}^{N H_F W_F \times 4}$, opacities $O \in \mathbb{R}^{N H_F W_F \times 1}$ and embeddings $E  \in \mathbb{R}^{N H_F W_F \times C}$. First, the module predicts a set of 3D gaussian for each camera in its own camera reference frame. Next, camera extrinsic parameters are applied to transform 3D gaussians from the camera to the world reference frame to finally concatenate all the gaussians into the single set $G$. 

The third module is the BeV rasterizer (Sec \ref{sec:bevrasterizer}) that performs a BeV rendering of the 3D gaussian set $G$ to produce the BeV feature map $B \in \mathbb{R}^{H_B \times W_B \times C}$, with $H_B$ and $W_B$ the dimensions of the BeV map.

Finally, in the last module, a BeV backbone and segmentation heads are sequentially applied to the BeV features to provide the final prediction.

\subsection{3D Gaussian generator}
\label{sec:gaussiangenerator}

Given the input feature maps $F$, the 3D gaussian generator predicts the 3D gaussian representation of the scene using several prediction heads. Figure \ref{fig:gg} illustrates how it operates on the feature maps. \\

\noindent\textbf{Gaussian centers.} The 3D positions of the gaussians in the scene are estimated by a depth head and a 3D offset head applied to $F$. The first predicts an initial position of the 3D centers along the optical rays. The second refines this 3D position by adding a small 3D displacement to it, giving more flexibility in the positioning of gaussians by not freezing them along the optical rays.

More precisely, for a pixel $i$ in the feature map of the camera $n$ with coordinates $(u_{n,i}, v_{n,i})$, the depth head predicts the disparity $d_{n,i} \in [0, 1]$ as in previous works dealing with monocular depth map estimation \cite{monodepth2, depthanything}. To compensate the influence of focal length diversity from one camera to another on depth prediction, disparity is predicted up to a scaling factor, in a reference focal length $f$ as proposed in \cite{metric3d}. Knowing the true focal length $f_n$ associated to the camera $n$, the metric depth $z_{n, i}$ is then decoded as follows:
\begin{equation}
z_{n,i} = \frac{f_n}{f}(\frac{1}{d_{n,i}} - 1)
\end{equation}
The corresponding 3D point $p^c_{n,i}$ in the camera reference frame is then deduced using the intrinsic matrix $K_n$ of the $n$-th camera: 
\begin{equation}
p^c_{n,i} = K_n^{-1}\begin{bmatrix} z_{n,i} u_{n,i} \\ z_{n,i} v_{n,i} \\ z_{n,i} \end{bmatrix}
\end{equation}

The resulting 3D points are constrained to lie along the optical ray passing through the pixel under consideration. Because of this constraint, their positioning is not necessarily optimal. To overcome this problem, we propose to use the 3D offset prediction head. It aims to provide a small displacement $\Delta_{n,i} = (\Delta x_{n,i}, \Delta y_{n,i}, \Delta z_{n,i})^T$ to be applied to the 3D center of the gaussian $p^c_{n,i}$ to refine its position in all three directions. The refined 3D point $p^c_{n,i}$ is simply obtained by:
\begin{equation}
\overline{p}^c_{n,i} = p^c_{n,i} + \Delta_{n,i}
\end{equation} 
At this stage, the 3D gaussian centers calculated for each camera are expressed in the corresponding camera reference frame. To express these points in the world reference frame, the extrinsic parameter matrices $[R_n | t_n]$ are applied, allowing the camera-to-world transformation :
\begin{equation}
p^w_{n,i} = [R_n | t_n]\overline{p}^c_{n,i}
\end{equation} 
The result is the gaussian center set $P  = \left\{ p^w_{n,i}\right\} \in \mathbb{R}^{N H_F W_F \times 3}$. \\ \\ 
\textbf{Gaussian rotations.}
The 3D rotations of the gaussians in the scene are estimated by a rotation head applied to $F$. For a given pixel in the feature map of the camera $n$, it outputs an allocentric rotation in the form of a unit quaternion $q^a_{n,i}$. The allocentric rotation of a pixel corresponds to a rotation relative to the 3D optical ray passing through it. This modelization makes it easier for the rotation head to learn, as it has no knowledge of the optical ray corresponding to the pixel it is processing. To take an example, two objects placed at two different locations in the scene and with different absolute (egocentric) rotations in the camera reference frame may have the same appearance in the image. In this case, the allocentric rotation predicted by the rotation head will be the same. The intrinsic parameters of the camera are then used to retrieve the egocentric rotation information.

For that purpose, the quaternion $q_{n, i}$ representing the rotation between the optical ray passing through the pixel $i$ of the camera $n$ and the axis $[0, 0, 1]^T$ is calculated. The quaternion $q^e_{n,i}$ representing the egocentric rotation in the camera reference frame is then recovered by:
\begin{equation}
q^e_{n,i} = q_{n, i} q^a_{n,i}
\end{equation} 
Finally, as for gaussian centers, the quaternion $q^w_{n,i}$ representing the rotation of the gaussian in the world reference frame is calculated using $q_{R_n}$, the quaternion modeling the camera-to-world rotation of the camera $n$:
\begin{equation}
q^w_{n,i} = q_{R_n} q^e_{n,i}
\end{equation} 
The quaternions thus calculated form the set of gaussian rotations $Q = \left\{ q^w_{n,i}\right\} \in \mathbb{R}^{N H_F W_F \times 4}$. \\

\noindent \textbf{Gaussian scales, opacities and features.} The last three gaussian parameters do not depend on optical properties and camera positioning, but rather encode semantic properties. Therefore, three heads are simply used to predict the sets, $S$, $O$ and $E$ required to render the gaussian set $G$ by the BeV rasterizer module.

\subsection{BeV rasterizer}
\label{sec:bevrasterizer}

The BeV rasterizer module is used to obtain the BeV feature map $B \in \mathbb{R}^{H_B \times W_B \times C}$ from the set of gaussians $G$ predicted by the 3D gaussian generator. To this end, the differentiable rasterization process proposed in gaussian splatting \cite{gs} has been adapted to perform this rendering. The first adaptation, already proposed in other offline semantic reconstruction works~\cite{langsplat}, consists in rendering $C$-dimensional features rather than colors. In our case, this produces a rendering containing semantic features essential for the perception task to solve. The second adaptation concerns the type of projection used. We have parameterized the rendering algorithm to generate orthographic rather than perspective renderings, more suited for BeV representation of the scene.

\subsection{GaussianBeV training}
\label{sec:losses}

 Our model is trained end-to-end using the loss functions used in \cite{fiery, simplebev, pointbev}.
 In these previous works, the semantic segmentation loss $L_{sem}$ is defined as follows:
\begin{equation}
L_{sem} = \lambda_{bce} L_{bce} + \lambda_{ctr} L_{ctr} + \lambda_{off} L_{off}
\label{eq:segm_loss}
\end{equation} 
With $L_{bce}$ the binary cross-entropy loss.
$L_{ctr}$ and $L_{off}$ correspond to the centerness and offset losses respectively, used as auxiliary losses to regularize training. $\lambda$ are weights to balance the three losses.

\paragraph*{Gaussian regularization losses.} Although GaussianBeV can be trained efficiently with the aforementioned losses, the addition of regularization functions acting directly on the gaussian representation improves its representational qualities. In particular, two regularization losses are added during training. 

First, a depth loss aims to regularize the position of the gaussians using the depth information provided by the projection of the LiDAR in the images. This loss adds constraints on the depth head predictions to obtain an initial 3D position, which is then refined by the 3D offsets (see Sec \ref{sec:gaussiangenerator}). With $z$ the ground truth depth and $z^*$ the predicted depth, the depth loss $L_{depth}$ is defined as follows:
\begin{equation}
L_{depth}(z, z^*) = \left| log(z) - log(z^*) \right|
\end{equation}
Second, an early supervision loss aims to optimize the gaussian representation before BeV backbone + heads are applied. The idea is to constrain the BeV features to directly provide all necessary information for the semantic segmentation task. In practice, segmentation heads are added and connected directly to the output of the BeV rasterizer module. The early supervision loss $L^{early}_{sem}$ is defined similarly to $L_{sem}$ (see equation \ref{eq:segm_loss}). The total loss function is therefore defined by:
\begin{equation}
L = L_{sem} + L^{early}_{sem} + \lambda_{depth} L_{depth},
\end{equation}
The influence of these learning strategies is analyzed in section \ref{sec:ablation}.

\section{Experiments}
\label{sec:expe} 

\begin{table*}[htb]
\centering
{\small
\begin{tabular}{lccccc}
\toprule
        & \multicolumn{1}{c}{}         & \multicolumn{2}{c}{all vehicles} & \multicolumn{2}{c}{vehicles with vis $> 40\%$} \\
        \cmidrule(lr){3-4} \cmidrule(l){5-6} 
 & \multicolumn{1}{c}{Image backbone} & 224$\times$448             & 448$\times$800             & 224$\times$448            & 448$\times$800           \\
\midrule
FIERY\cite{fiery} & EN-b4 & 35.8 & - & 39.8 & -\\
CVT \cite{cvt}      & EN-b4 & 31.4 & 32.5 & 36.0 & 37.7\\
LaRa \cite{lara}     & EN-b4 & 35.4 & - & 38.9 & - \\
BEVFormer \cite{bevformer}      & RN-50 & 35.8 & 39.0 & 42.0 & 45.5 \\
Simple-BEV \cite{simplebev}      & RN-50 & 36.9 & 40.9 & 43.0 & 46.6\\
BAEFormer \cite{beaformer}      & EN-b4 & 36.0 & 37.8 & 38.9 & 41.0\\
PointBeV \cite{pointbev}     & EN-b4 & 38.7 & 42.1 & 44.0 & 47.6\\
\midrule
GaussianBeV    & EN-b4 & \textbf{41.6} & \textbf{43.9} & \textbf{47.5} & \textbf{50.3}                  \\
\bottomrule
\end{tabular}
}
\caption{GaussianBeV IoU results on vehicle segmentation on the nuScenes validation set using different input resolutions and different visibility filtering. Our method remarkably outperforms previous works on all experimental settings.}
\label{tab:vehicle}
\end{table*}

\noindent \textbf{Dataset.} We use the nuScenes dataset  \cite{nuscenes} consisting in a set of multiview sequences obtained with a system of $N$=6 surrounding cameras. It is divided into 750 sequences (28,130 images) for training and 150 sequences (6,019 images) for validation. It provides 3D bounding box annotations (vehicles and pedestrians) and semantized HDmap annotations of the road surface (drivable area and lane boundaries). \\
\noindent \textbf{Architecture details.} For image feature extraction, the EfficientNet-b4 backbone \cite{effnet} and the Simple-BeV neck \cite{simplebev} are used. For the BeV backbone, unless specified, the model of LSS \cite{lss} and the segmentation heads of Simple-BEV \cite{simplebev} are connected to the output of the BeV rasterizer. We use a channel size of $C$=128 for all experiments. For gaussian prediction heads, we use a light convolutional neural network composed by two blocks of Conv-Batchnorm-ReLU followed by the output layer. The sigmoid activation function is applied on both disparity and opacity outputs. L2 normalization and absolute value activation are applied on the rotation and scale outputs, respectively.    \\
\noindent \textbf{Implementation details.} Following previous work \cite{pointbev}, GaussianBeV is trained on a maximum of 100 epochs using an AdamW optimizer with learning rate of 3e-4, a weight decay of 1e-7 and a one-cycle linear learning rate scheduler.
As in \cite{fiery, pointbev}, loss functions are balanced using uncertainty weighting as proposed in \cite{kendall}.
We use random scaling, random crop and random flip for images augmentation and random translation and rotation for BeV augmentation.
Two input image resolutions $H$$\times$$W$ are tested in the experiments, 224$\times$480 and 448$\times$800. The resulting image feature maps have a size $H_F$$\times$$W_F$ of 28$\times$60 and 56$\times$100, respectively.
GaussianBeV training on the lower resolution uses a batch size of 11 on an A100 GPU and for the higher resolution a batch size of 9, distributed over 3 A100 GPUs.
The BeV rasterizer output a BeV feature map of size 200 x 200 representing a 100m×100m BeV representation with a 50cm resolution.
The IoU metric is used on the validation set for evaluation.

\subsection{State-of-the-art comparison}

\noindent \textbf{Vehicle segmentation.} We compare GaussianBeV with previous works on vehicle semantic segmentation using different input resolutions (224$\times$448 and 448$\times$800) and different visibility filtering :  (1) considering all vehicles and (2) only keeping vehicles with visibility $> 40\%$. Results are given in Table \ref{tab:vehicle}. It shows that GaussianBeV clearly outperforms previous methods on all experimental settings. For instance, it supasses the previous state-of-the-art method PointBeV \cite{pointbev} by +3.5 IoU for the experiments using 224$\times$448 input resolution and visibility filtering. 

\noindent \textbf{Pedestrian segmentation.} We also compare GaussianBeV to other previous methods on the pedestrian segmentation task. For this evaluation, we use the 224$\times$448 input resolution and visibility filtering. Results are available in Table~\ref{tab:pedestrian}. Once again, GaussianBeV outperforms previous state-of-the-art method PointBeV\cite{pointbev} by +2.7 IoU. 

\noindent \textbf{Ground surface segmentation.} We train GaussianBeV for joint segmentation of drivable area and lane boundaries using an input resolution of 448$\times$800. Results are given in Table~\ref{tab:map}. Compared to previous state-of-the art method MatrixVT\cite{matrixvt}, GaussianBeV gives superior results for lane boundary segmentation (+2.6 IoU). However, our method is slightly less effective in segmenting the drivable area (-0.9 IoU) than MatrixVT. The ability of GaussianBeV to model the scene in details enables better segmentation of fine structures, but does not improve performance in larger areas that are easier to segment. 

\noindent \textbf{Inference time.} We compared for both input resolutions the inference time of GaussianBeV and the previous method, PointBeV \cite{pointbev}, on an A100 GPU. GaussianBeV runs at 24 fps and 13 fps while PointBeV runs at 19 fps and 15 fps. Our method is therefore comparable in terms of computation time, but future research may enable to speed up the model by optimizing the gaussian representation. 

\begin{table}[htb]
\centering
{\small
\begin{tabular}{lc}
\toprule
     & Pedestrians   \\
\midrule
LSS \cite{lss}   &    15.0    \\
FIERY \cite{fiery}    &    17.2   \\
ST-P3  \cite{stp3}   &    14.5    \\
TBP-Former static \cite{tbp}   &    17.2   \\ 
PointBEV \cite{pointbev}   &    18.5     \\
\midrule
GaussianBeV &       \textbf{21.2}    \\
\bottomrule
\end{tabular}
}
\caption{GaussianBeV IoU results on pedestrian segmentation on the nuScenes validation set. GaussianBeV is trained with 224$\times$448 input image resolution and visibility filtering. Our method increases performances compared to previous works.}
\label{tab:pedestrian}
\end{table}

\begin{table}[]
\centering
{\small
\begin{tabular}{lcc}
\toprule
 & Drivable area & Lane boundaries \\
\midrule
LSS \cite{lss}  &      72.9         & 20.0               \\
M$^2$BEV \cite{m2bev} &  75.9             & 38.0               \\
BEVFormer \cite{bevformer}  & 80.1              & 25.7               \\
PETRv2 \cite{petrv2}  &        83.3       & 44.8               \\
MatrixVT \cite{matrixvt}  &    \textbf{83.5}           & 44.8 \\
PointBeV \cite{pointbev}$^{\dagger}$ & 72.5 & 37.6 \\
\midrule
GaussianBeV  &      82.6         &     \textbf{47.4}           \\
\bottomrule
\end{tabular}
}
\caption{GaussianBeV IoU results on both drivable area and lane boundary segmentation on the nuScenes validation set. GaussianBeV is trained with 448$\times$800 input image resolution. ${\dagger}$ means that we trained the model by ourselves. Our method outperfoms previous works on lane boundary segmentation and gives competitive results on drivable area segmentation.}
\label{tab:map}
\end{table}

\begin{figure*}[htbp]
  \centering
  \includegraphics[width=0.85\linewidth]{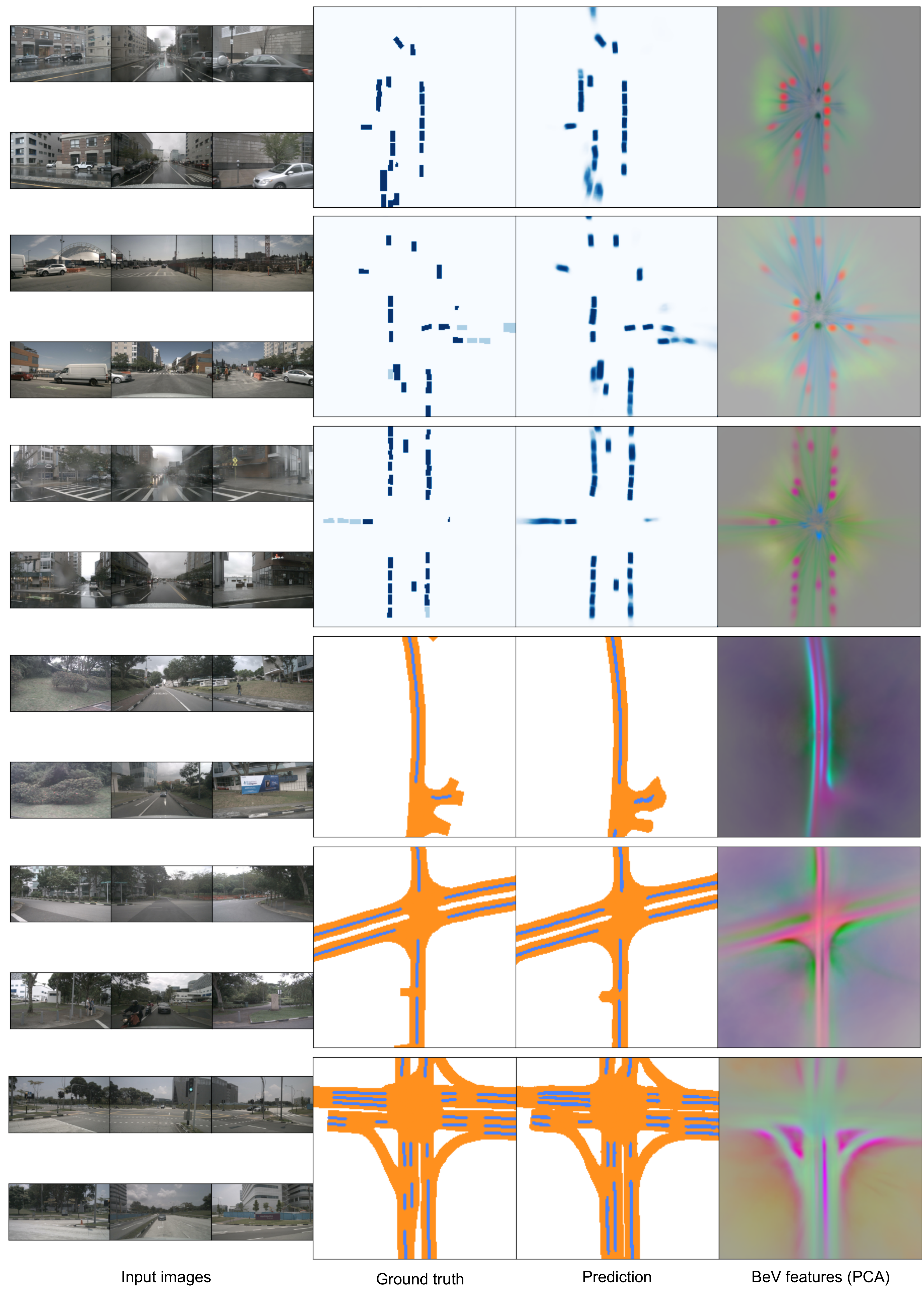}
  \caption{Visualization of predicted vehicle segmentation in the first three rows and drivable area (blue) / lane boundary (orange) segmentation in the last three rows, on the nuScenes validation set. PCA is used to vizualize the BeV feature map.}
  \label{fig:visuall}
\end{figure*}

\subsection{Ablations}
\label{sec:ablation}

\begin{table}[]
\centering
{\small
\begin{tabular}{cccccc}
\toprule
Depth sup & Early sup & BeV backbone & rot & off &  IoU \\ 
\midrule
         &   \ding{51}       &  none & \ding{51} & \ding{51} & 45.4\\
  \ding{51}      &   \ding{51} &  none & \ding{51} & \ding{51} & 45.9\\
\midrule
          & & LSS \cite{lss} & \ding{51} & \ding{51} & 47.0 \\
  \ding{51}      &         & LSS \cite{lss} & \ding{51} & \ding{51} & 46.9 \\
        &    \ding{51}     & LSS \cite{lss} & \ding{51} & \ding{51} & 46.7 \\
\midrule
   \ding{51}   &   \ding{51}  & LSS \cite{lss} &  &  & 47.1 \\
   \ding{51}   &    \ding{51} & LSS \cite{lss} & \ding{51} &  & 47.3 \\
  \ding{51}    &    \ding{51} & LSS \cite{lss} &  & \ding{51} & 47.1 \\
\midrule
  \ding{51}      &    \ding{51}     & SimpleBeV \cite{simplebev} & \ding{51} & \ding{51} & 47.0 \\
   \ding{51}      &    \ding{51}     &LSS \cite{lss} & \ding{51} & \ding{51} & \textbf{47.5}\\
\bottomrule
\end{tabular}
}
\caption{Influence of BeV backbone, depth and early supervision, rotation and 3D offset prediction. Models are trained using 224$\times$448 input resolution and visibility filtering on the vehicle segmentation task.}
\label{tab:ablation}
\end{table}

\begin{figure}[ht]
    \centering
    \begin{subfigure}[b]{0.13\textwidth}
        \includegraphics[width=\textwidth]{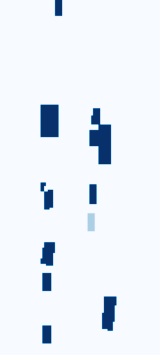}
        \caption{Ground truth}
        \label{fig:gtearly}
    \end{subfigure}
    \begin{subfigure}[b]{0.13\textwidth}
        \includegraphics[width=\textwidth]{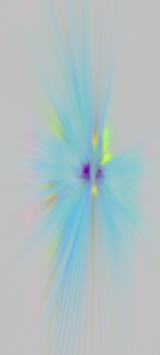}
        \caption{w/o early sup}
        \label{fig:nearly}
    \end{subfigure}
    \begin{subfigure}[b]{0.13\textwidth}
        \includegraphics[width=\textwidth]{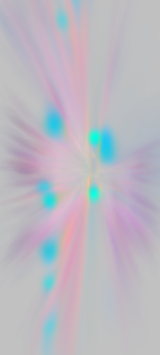}
        \caption{w/ early sup}
        \label{fig:early}
    \end{subfigure}
    \caption{Influence of the early supervision. (a) Vehicle segmentation ground truth. (b) and (c) visualization of the BeV feature map for GaussianBeV trained without and with early supervision, respectively. Early supervision helps to achieve a representation closer to the content of the 3D scene. We observe that gaussians corresponding to the vehicles follow their shapes.}
    \label{fig:comp}
\end{figure}

Table \ref{tab:ablation} illustrates the influence of different design choices and training strategies. The Figure \ref{fig:visuall} shows qualitative results for vehicle and ground surface segmentation. It also gives a visualization of the output of the BeV rasterizer module. To visualize the BeV feature map, we applied Principal Component Analysis (PCA) to the 3D gaussian features before BeV rendering. This allows us to reduce the size of the features to render the BeV features in color.  \\

\noindent \textbf{BeV backbone.} We conducted experiments to evaluate the influence of the choice of the BeV backbone. In particular, GaussianBeV is trained (1) without using a BeV backbone, by connecting the BeV feature directly to the segmentation heads (row 2 in the Table \ref{tab:ablation}) and (2) with the Simple-BeV backbone \cite{simplebev} to compare performance with the standard LSS backbone \cite{lss} (rows 9 and 10 in the Table \ref{tab:ablation}). Remarkably, GaussianBeV trained without BeV backbone is already yielding good results (45.9 IoU that is effectively superior to previous methods). This shows that GaussianBeV is able to provide a geometrically and semantically relevant gaussian representation that can be directly exploited for segmentation. With this observation, the BeV backbone is used to refine the representation using multi-scale information. We observe that the BeV backbone from SimpleBeV \cite{simplebev} achieves very good results (47.0 IoU), even if they are slightly inferior to those obtained with the BeV backbone from LSS \cite{lss} (47.5 IoU). 

\noindent \textbf{Depth and early supervision.} We investigate the influence of auxiliary losses on the gaussian representation as detailed in Sec \ref{sec:losses}. As shown in rows 1 and 2 of the Table \ref{tab:ablation}, the use of the depth loss increases performance when GaussianBeV is trained without a BeV backbone. This shows that this loss regularizes the model to help it outputs a more geometrically consistent gaussian representation. Note that early supervision is used by default if the BeV backbone is not used, as this becomes the main segmentation loss. With regard to rows 3, 4 and 5 of the table using the entire BeV backbone, depth supervision and early supervision do not bring any performance gains when used separately. However, the last line of the table shows that, used together, these two losses improve results. In Figure \ref{fig:comp}, we provide a visualization of the impact of early supervision on the 3D gaussian representation. We observe that using early supervision results in a more coherent BeV feature map. Indeed, the gaussians better fit the shape of the vehicles. 

\noindent \textbf{3D offsets and rotations.} Table \ref{tab:ablation} (rows 6 to 8) shows the impact of learning 3D offsets and rotation. When GaussianBeV is trained without these two heads, it gives slightly poorer results than when it is learned with (last row). 

\section{Conclusion}
\label{sec:conclusion}

In this paper, we introduced GaussianBeV, a novel image-to-BeV transformation method that is the new state-of-the-art on BeV semantic segmentation.
Based on an \textit{optimization free} 3D gaussian generator, it transforms each pixel of an image feature map into semantized 3D gaussians.
Gaussians are then splattered to obtain the BeV feature map. We have shown that the gaussian representation enables fine 3D modeling thanks to its ability to adapt to the different geometric structures present in the scene. Future work will aim to generate a more compact 3D Gaussian representation to improve computing time. We hope that this initial work will open the door to further research in 3D perception using \textit{optimization free} gaussian splatting representation. \\

\noindent \textbf{Acknowledgements.} This work was carried out as part of the CEA-Valeo joint laboratory and partially financed by BPI France as part of the AVFS project. This work benefited from the FactoryIA supercomputer financially supported by the Ile-de-France Regional Council.

{\small
\bibliographystyle{ieee_fullname}
\bibliography{main}
}

\end{document}